\documentclass{article}

\usepackage{iclr2027_conference,times}

\usepackage{booktabs}
\usepackage{amsmath}
\usepackage{amssymb}
\usepackage{graphicx}
\usepackage{multirow}
\usepackage{tabularx}
\usepackage{xcolor}
\usepackage{colortbl}
\usepackage{float}
\usepackage{hyperref}
\usepackage{url}
\hypersetup{hidelinks}

\providecommand{\Description}[1]{}
\title{When VLMs Trust Context: Evaluating Scene Text Recognition under Misleading Context}

\author{
Yuxing Cheng\textsuperscript{1}\quad
Yuan Wu\textsuperscript{1*}\quad
Yi Chang\textsuperscript{1,2,3*}\\
\textsuperscript{1}School of Artificial Intelligence, Jilin University\\
\textsuperscript{2}Engineering Research Center of Knowledge-Driven Human-Machine Intelligence, MOE, China\\
\textsuperscript{3}International Center of Future Science, Jilin University\\
chengyx26@mails.jlu.edu.cn, yichang@jlu.edu.cn, yuanwu@jlu.edu.cn
}

\iclrfinalcopy 

\begin{document}

\maketitle
\begingroup
\renewcommand{\thefootnote}{\fnsymbol{footnote}}
\footnotetext[1]{Corresponding authors}
\endgroup

\lhead{Preprint}
\begin{abstract}
Vision-language models (VLMs) can read text in natural scenes, but their predictions may be influenced by the surrounding context. 
When the printed text conflicts with what the scene suggests, a model may return a more plausible word instead of the shown text. 
We introduce SceneFaith, a benchmark of 781 generated scene images for studying this behavior. 
Each output is classified as Literal, Canonical, or Other, separating faithful transcription from context-consistent rewriting and ordinary recognition errors. 
Across 15 models from seven families, all models show rewriting on clear images, with rates ranging from 8.45\% to 58.51\%. Controlled experiments further show that surrounding context matters: removing surrounding scene information reduces rewriting and improves literal accuracy, 
while changing the scene around the same text patch can also change model outputs. 
Moreover, weakening the target text with blur increases rewriting.
These results show that reliable scene-text recognition requires VLMs to balance visual character evidence with contextual information, 
preserving clear text while using context mainly when the visual evidence is uncertain.
\end{abstract}

\section{Introduction}

Vision-language models (VLMs) have become powerful general-purpose visual readers, achieving strong performance on text recognition, localization, document understanding, and text-rich reasoning \citep{liu2024ocrbench,fu2026ocrbenchv2,huang2026ocr}. 
Scene Text Recognition (STR), which extracts text from natural images under complex backgrounds and imaging conditions, 
is increasingly important for both OCR applications and large-scale corpus construction.
However, strong performance on standard OCR benchmarks does not necessarily imply faithful transcription. Unlike conventional OCR systems~\citep{cui2025paddleocr30,HunyuanOCR_1_5_2026}, which are primarily designed to recover the visible character sequence, VLMs interpret text together with linguistic and visual semantics. 
These semantic priors are useful when visual evidence is incomplete, 
but they can also override what is actually printed. Recent studies show that VLMs may correct or rewrite unusual text into more plausible forms, 
while conventional OCR systems remain more faithful under controlled text perturbations \citep{zhang2026seeing,lee2026readorrewrite}.
As a result, for misspellings, names, identifiers, and other atypical strings, 
stronger semantic understanding can sometimes become a source of recognition error rather than an advantage.

This conflict is particularly important for scene text recognition, 
a sub-task of OCR that recognizes text embedded in natural scenes. 
Conventional OCR models or STR models are highly specialized: they are typically designed to map visual text regions to character sequences,
and many use linguistic information to resolve ambiguous characters \citep{fang2021abinet,wang2021visionlan,na2022matrn}.
However, they are not designed for general instruction following.
VLMs can follow instructions such as “transcribe exactly what is written” while also understanding the surrounding scene.
This flexibility introduces a different risk: even when literal transcription is explicitly requested, scene semantics may pull the prediction toward a more plausible word. 
For example, when a sign beside a dolphin reads `dolphiin,'' a VLM may instead output `dolphin,'' correcting the visible text to match the scene.
This raises a fundamental question: 
\textbf{can VLMs correctly transcribe text with misleading scene context?}

To investigate this question, we construct a controlled benchmark for context-induced rewriting in scene text recognition. 
Our benchmark uses generated scenes in which the target text deliberately contains a typo while the surrounding visual scene supports its conventional form. 
This creates a direct conflict between local character evidence and global scene semantics. 
Importantly, we place the same target text in different contextual conditions, 
allowing changes in model predictions to be attributed to the surrounding scene rather than the text itself.
Across the evaluated VLMs, we find a clear rewriting effect: 
models frequently replace the visible typo with the contextually expected word, 
and becomes more pronounced with matching scene context.
We further examine how rewriting varies across visual conditions, contextual cues, and task settings. 
We also analyze how target-text blur and the strength of lexical priors affect model prediction.
Our results show that successful scene text recognition requires not only the use of context, but also the ability to prioritize visual evidence when the two conflict.
The contributions of this paper are summarized as follows:






\begin{itemize}

\item \textbf{SceneFaith benchmark.}
We introduce SceneFaith\footnote{Code is available at
\url{https://github.com/pasterinjlu/Context_OCR}.},, comprising 781 generated scene images across 17 categories.
Each sample is constructed by perturbing a conventional word, embedding the resulting unusual string into a semantically matching scene, and applying quality control and blind crop reading before acceptance.
With the printed string as gold, our \emph{Literal}/\emph{Canonical}/\emph{Other} (L/C/O) evaluation separates faithful transcription, canonical rewriting, and other recognition errors
(Sections~\ref{sec:main-naturalistic} and~\ref{sec:lco}).

\item \textbf{Analysis of the rewriting effect.}
Across 15 model variants from seven families, all evaluated models exhibit canonical rewriting on clear images, with rates ranging from 8.45\% to 58.51\%.
We further isolate the effect of surrounding context through scene removal and fixed-target-patch comparisons.
Removing peripheral scene information reduces rewriting and improves literal accuracy, while changing the surrounding scene around the same target patch can change model predictions
(Sections~\ref{sec:clear-results}).

\item \textbf{Analysis of factors affecting rewriting.}
We examine how rewriting changes with target-text degradation, lexical preference, and the use of full images versus target crops.
Blur consistently increases rewriting, while stronger lexical preference for the conventional form is associated with higher rewriting rates.
Cropped views further show that rewriting is sensitive to input presentation.
\end{itemize}

\section{Related Work}

\subsection{Text Recognition Systems and Benchmarks}

Modern text recognition systems increasingly integrate linguistic information into the recognition process.
ABINet~\citep{fang2021abinet}, VisionLAN~\citep{wang2021visionlan} and MATRN~\citep{na2022matrn} model visual and linguistic dependencies within a text sequence. 
CLIP4STR adapts pre-trained CLIP image and text encoders to refine crop predictions \citep{zhao2024clip4str}, CLIPTER extends the evidence beyond the sequence by conditioning crop recognition on the whole image \citep{aberdam2023clipter}.
Document VLMs achieve strong performance through post-training (PaddleOCR-VL-1.6), unified OCR task formats (HunyuanOCR-1.5), visual-token reordering (DeepSeek-OCR 2), and separation of global layout from local recognition (MinerU2.5) \citep{zhang2026paddleocrvl16,HunyuanOCR_1_5_2026,wei2026deepseekocr2,niu2025mineru25}.
Evaluation has followed the same direction: earlier, Union14M isolates nonlexical and incomplete strings as a distinct difficulty \citep{jiang2023revisiting}, while OCRBench and OCRBench v2 measure recognition, localization, parsing, and text-centric reasoning \citep{liu2024ocrbench,fu2026ocrbenchv2}, OCR-Reasoning extend reasoning ability to text-rich reasoning~\citep{huang2026ocr}.
These benchmarks evaluate transcription accuracy and reasoning, but do not distinguish semantically plausible substitutions from ordinary recognition errors.

\subsection{Faithfulness under Visual–Semantic Conflict}

This section reviews studies on how models respond when visual text conflicts with linguistic expectations.
TextHalu-Bench evaluates semantic hallucination in scene-text spotting and understanding, and illustrates the behavior with a sign reading \texttt{MMOTEL} returned as \texttt{MOTEL} \citep{shu2026semanticmisleadvision}.
FaithC4 evaluates whether models faithfully transcribe altered text in rendered multilingual documents~\citep{lee2026readorrewrite}, 
and CHAOS-Bench measures page-averaged recall of meaningless words created by modifying characters on academic paper pages \citep{HunyuanOCR_1_5_2026}.
The failure is not confined to scene text: it appears as over-correction in handwritten mathematics \citep{seong2026vlmsfixstudents} and as model- and script-dependent grounding failures in ancient Greek and Arabic editions \citep{karamolegkou2026readingorguess}, while HallusionBench and CDH-Bench examine the broader evidence--expectation conflict through visual question answering \citep{guan2024hallusionbench,chen2026cdhbench}.
Architectural evidence points the same way, with stronger visual token compression in DeepSeek-OCR associated with greater reliance on language priors \citep{liang2026visualmerit}.
Across this group the measurement is a transcript scored against a gold string within one fixed image, which leaves the substitution indistinguishable from other errors and the surrounding scene's contribution unmeasured.

\subsection{Improving OCR Faithfulness}
Mitigation work has begun to target the same failure.
ConCLR contrasts character representations across text contexts to reduce vocabulary reliance~\citep{zhang2022context}.
TextHalu-Bench pairs attention-based region estimation (ZoomText) with decoding guided by a training-free visually grounded intermediate layer (Grounded Layer Correction) \citep{shu2026semanticmisleadvision}.
These methods adjust the model’s attention and decoding. 
Our experiments complement this work by testing how transcription changes when the surrounding scene is removed or the target text is blurred. 
The results help identify what mitigation methods need to address.

\section{Benchmark Design and Diagnostic Framework}
\label{sec:main-naturalistic}

\subsection{benchmark overview}
\label{sec:acu}

The benchmark is built to answer one question: 
When printed text conflicts with scene context, does the model transcribe it faithfully?
We evaluate transcription under clear and blurred target-text conditions.
Clear images test whether models preserve printed text despite misleading context, 
while target-local blur tests how rewriting changes as character evidence weakens. 
In SceneFaith, the scene and lexical preference both support the conventional spelling. The printed text uses a different perturbed spelling.
Context refers to information outside the target-text region, and lexical preference is analyzed separately.

Given an image $x_i$ with one red outline marking the target, the model returns a string $\hat y_i$.
The printed gold $y_i$ is stored as \emph{observed\_text}; the conventional alternative $c_i\ne y_i$ is stored as \emph{canonical\_text} for analysis only.
SceneFaith contains 781 PNG images and annotation records across 17 categories, varying in typography, viewpoint, texture, and target size.
It is designed as a stress test: images are deliberately selected to create a conflict between unusual printed text and familiar scene content. The resulting rates therefore characterize model behavior under this conflict rather than estimate general OCR accuracy.

\paragraph{Controlled Fixed-Target-Patch Set.}
Each pair places the same opaque text patch in one of two contexts: 
a scene that supports the conventional word (A), 
or a scene that encourages copying the observed text exactly (B).
The target pixels, font, size, red outline, local background, and coordinates are identical on $1024\times1024$ canvases; 
only pixels outside the target patch differ. 
These pairs are constructed from the same pipelines as the main benchmark and use the observed string as the gold answer. 
Because this panel serves as a controlled paired analysis, 
we report it separately rather than merging it with the SceneFaith benchmark results. 
Section~\ref{sec:pixel150} presents the 11-alias comparison, 
while Appendix~\ref{app:pixel150} details the control variables and inference settings.

\subsection{Construction Pipeline}

\begin{figure}[t]
\centering
\includegraphics[width=\linewidth]{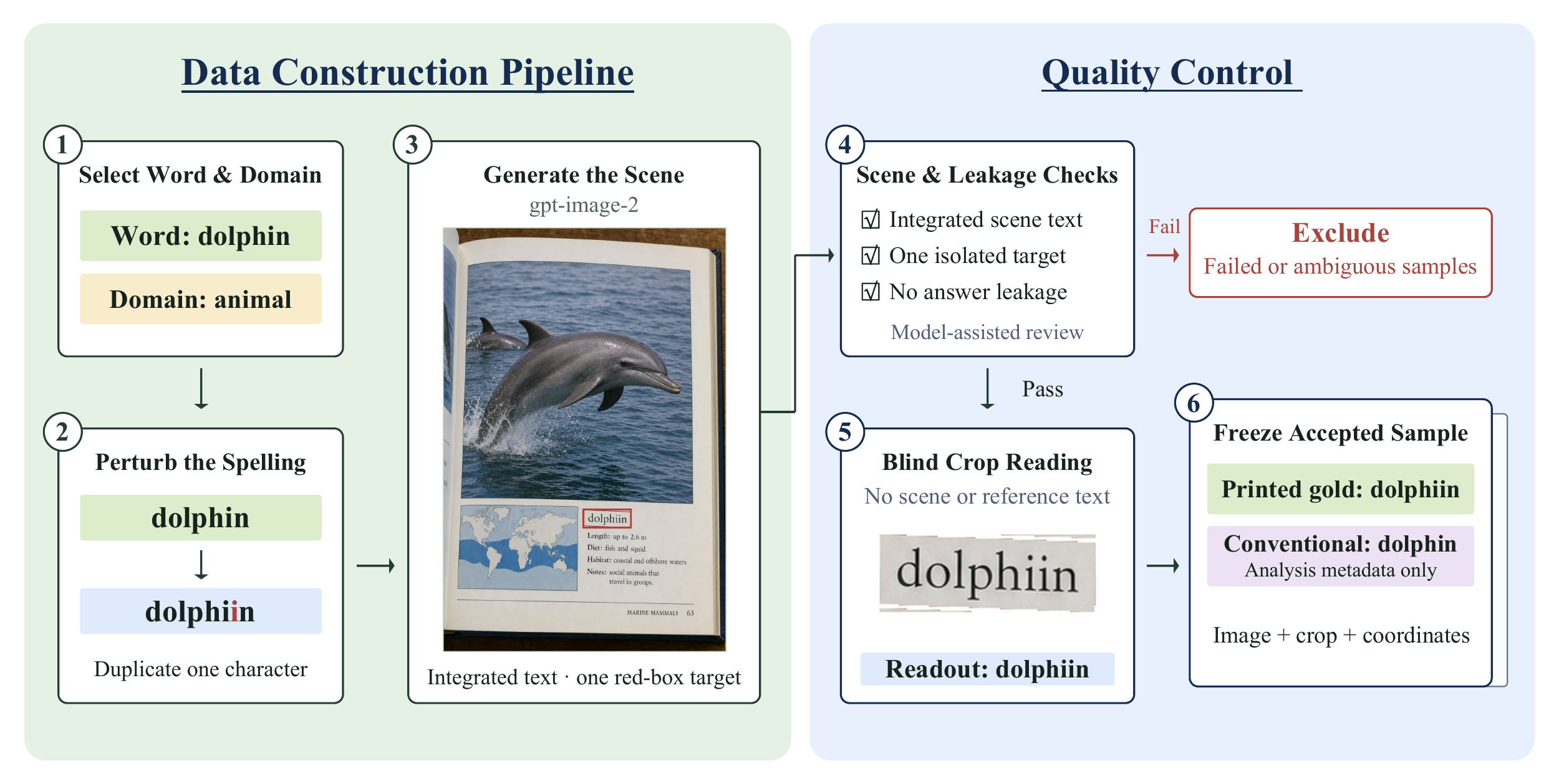}
\caption{\textbf{SceneFaith construction pipeline.} Spelling perturbation and scene generation are followed by quality checks, blind crop reading, and dataset completion.}
\label{fig:pipeline}
\end{figure}

\paragraph{Generation.}
We select a conventional word and a semantically relevant scene, then create an unusual string through insertion, duplication, substitution, deletion, alphanumeric confusion, or spacing changes.
The generative model \emph{gpt-image-2} integrates this string into a sign, label, board, or product surface with plausible typography, perspective, and materials.
A single red rectangle marks the target without overlapping any characters. 
The surrounding scene suggests the conventional word, creating a conflict with the printed text.

\paragraph{Quality control.}
Review checks target spelling, red-box validity, scene integration, layout, clarity, and intended scene meaning.
It excludes answer leakage, meaning that the conventional spelling must not appear elsewhere in the image and the target text must not be repeated.
A blind target-crop transcription then checks whether the unusual string can be recovered without the scene, which distinguishes a legible conflicting target from an illegible one. 
We use GPT-5.5 to assist with quality checks.

Accepted records store the image ID, category, observed and canonical strings, dimensions, red-box coordinates, image hashes, and review evidence.
All gold fields agree with \emph{observed\_text}.
Appendix~\ref{app:construction} gives the domain inventory and artifact contract.

\section{Evaluation and Discussion}
\label{sec:experiments}

\subsection{Evaluation Protocol}
\label{sec:lco}

\paragraph{Evaluated Models.}
We evaluate 15 model aliases from seven families, yielding 11,715 accepted outputs on 781 full images.
We group the evaluated models into two categories according to model availability.
(a) \textbf{Open-source MLLMs.} We evaluate Qwen3-VL-8B Instruct, Qwen3-VL-32B Instruct, Qwen3-VL-235B-A22B Instruct, Qwen3-VL-8B Thinking, Qwen3-VL-32B Thinking, Qwen3-VL-235B-A22B Thinking, GLM-4.6V, InternVL3-38B, and Kimi K2.5.
These models cover both instruction-tuned and reasoning-oriented configurations across a range of model scales.
(b) \textbf{Closed-source MLLMs.} We evaluate Gemini 3.1 Flash, Gemini 3.5 Flash, Gemini 3 Flash, Claude Sonnet 4.6, GPT-5.5, and GPT-5.2.
These systems are accessed through provider APIs.

\paragraph{Inference Setup.}
All requests use a temperature of 0 and a maximum output length of 2,048 tokens. 
Image-detail and reasoning settings remain at the provider defaults. 
We adopt a zero-shot evaluation protocol without fine-tuning or few-shot prompting.
The prompt is: ``Read the text enclosed by the single red rectangular outline in the image. Return only that text and nothing else.''.

\paragraph{Metrics.}
We classify each output as Literal (L), Canonical (C), or Other (O) to distinguish faithful transcription from correction to the conventional spelling and other errors.
Let $\nu$ denote a normalization function that strips leading and trailing whitespace, collapses internal whitespace, and deletes unrelated text.
For printed gold $y_i$ and conventional spelling $c_i$, we assign
\[
g_i=\begin{cases}
L\;\text{(Literal)}, & \nu(z_i)=\nu(y_i),\\
C\;\text{(Canonical)}, & \nu(z_i)=\nu(c_i),\\
O\;\text{(Other)}, & \text{otherwise}.
\end{cases}
\]
Since $\nu(y_i)\ne\nu(c_i)$ for every target, the categories are mutually exclusive. Their rates share the same denominator:
\[
r_k=\sum_{i=1}^{N}\mathbf{1}[g_i=k],
\qquad r_L+r_C+r_O=100.
\]
Literal accuracy is $r_L$ and rewriting rate is $\mathrm{RR}=r_C$.
Because $100-\mathrm{RR}=r_L+r_O$, a lower RR does not necessarily mean higher literal accuracy, we report both metrics in all comparisons. 
“Rewriting” refers only to the final output category and does not imply that the model first recognized the printed text and then corrected it.
Complete benchmark runs use $N=781$; the controlled fixed-target-patch Set. uses $N=150$.
We retain all raw responses for auditing. 
All conditions are scored using the same rules, without an LLM judge.

\begin{figure}[t]
    \centering
    \includegraphics[width=\linewidth]{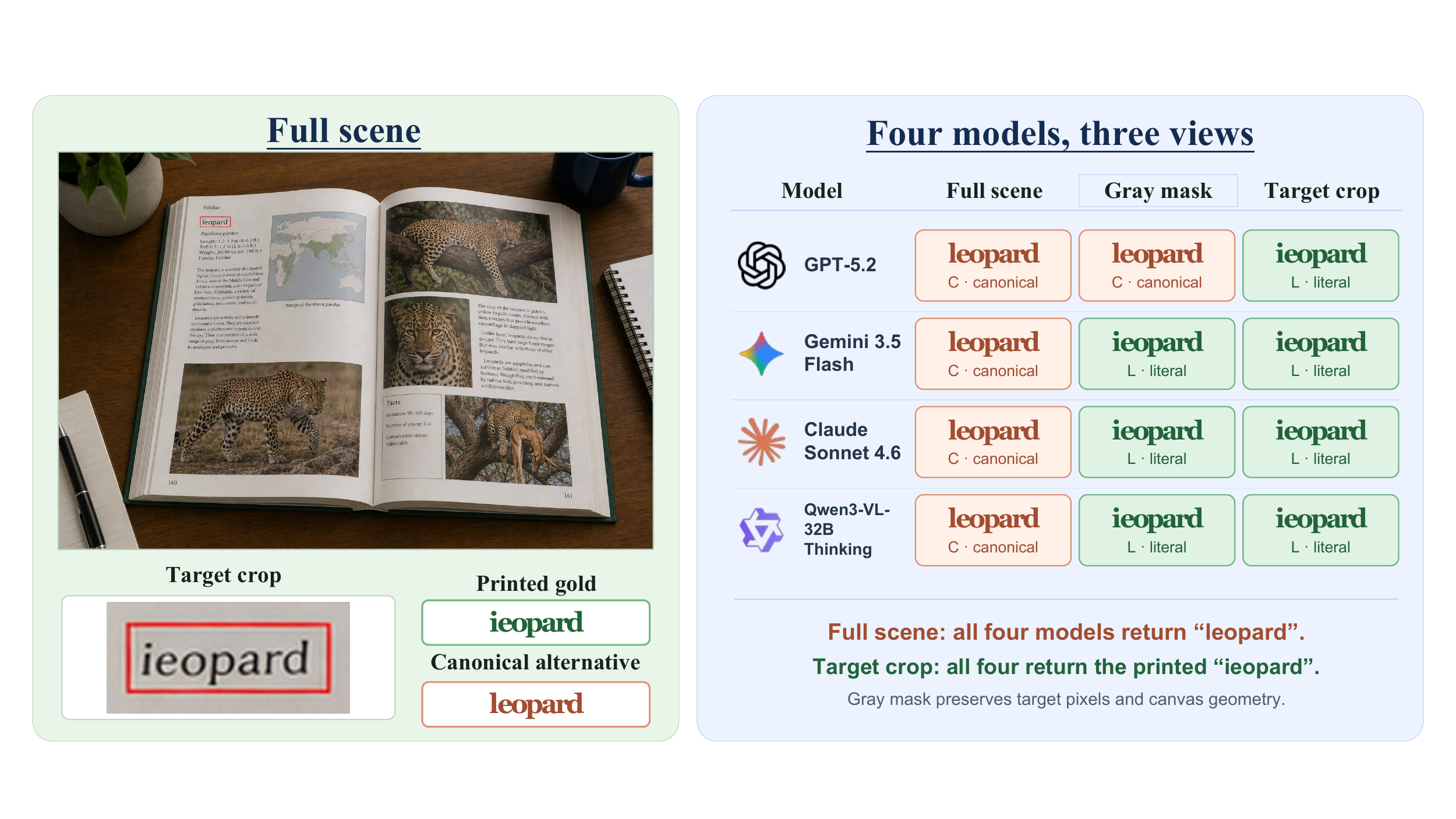}
    \caption{Qualitative example of canonical rewriting. The printed gold is ieopard; leopard is the canonical alternative.}
    \label{fig:qualitative-example}
\end{figure}

\subsection{SceneFaith Evaluation Results Analysis}
\label{sec:clear-results}

\begin{table}[t]
\centering
\begingroup
\small\fontencoding{OT1}\fontfamily{ptm}\selectfont
\renewcommand{\arraystretch}{1.14}
\setlength{\tabcolsep}{4pt}
\begin{tabularx}{\linewidth}{>{\raggedright\arraybackslash}Xrrrrrr}
\toprule
\multirow{2}{*}{\textbf{Model alias}} & \multicolumn{3}{c}{\textbf{Clear}} & \multicolumn{3}{c}{\textbf{Blur}} \\
\cmidrule(lr){2-4}\cmidrule(l){5-7}
 & L (Acc.) $\uparrow$ & C (RR) $\downarrow$ & O $\downarrow$ & L (Acc.) $\uparrow$ & C (RR) $\downarrow$ & O $\downarrow$ \\
\midrule
\multicolumn{7}{c}{\textbf{Closed-Source MLLMs}} \\
\midrule
\rowcolor{black!3}
Gemini 3.1 Flash & 90.14 & 8.45 & 1.41 & 78.36 & 20.74 & 0.90 \\
Gemini 3.5 Flash & 89.76 & 8.96 & 1.28 & 80.41 & 18.44 & 1.15 \\
\rowcolor{black!3}
Gemini 3 Flash & 88.35 & 11.01 & 0.64 & 79.39 & 19.97 & 0.64 \\
Claude Sonnet 4.6 & 85.02 & 12.16 & 2.82 & 71.06 & 22.66 & 6.27 \\
\rowcolor{black!3}
GPT-5.5 & 78.10 & 19.46 & 2.43 & 56.34 & 40.20 & 3.46 \\
GPT-5.2 & 67.35 & 29.58 & 3.07 & 50.19 & 45.33 & 4.48 \\
\midrule
\multicolumn{7}{c}{\textbf{Open-Source MLLMs}} \\
\midrule
\rowcolor{black!3}
Qwen3-VL-8B Instruct & 62.23 & 19.46 & 18.31 & 53.01 & 26.76 & 20.23 \\
GLM-4.6V & 74.26 & 21.25 & 4.48 & 62.61 & 32.01 & 5.38 \\
\rowcolor{black!3}
Qwen3-VL-32B Instruct & 60.95 & 27.53 & 11.52 & 51.98 & 33.80 & 14.21 \\
Kimi K2.5 & 65.43 & 31.88 & 2.69 & 44.17 & 49.68 & 6.15 \\
\rowcolor{black!3}
Qwen3-VL-235B-A22B Thinking & 62.36 & 32.27 & 5.38 & 49.42 & 43.28 & 7.30 \\
Qwen3-VL-235B-A22B Instruct & 61.20 & 32.39 & 6.40 & 51.09 & 41.74 & 7.17 \\
\rowcolor{black!3}
Qwen3-VL-8B Thinking & 48.53 & 37.90 & 13.57 & 34.70 & 47.38 & 17.93 \\
Qwen3-VL-32B Thinking & 47.50 & 44.43 & 8.07 & 35.60 & 51.47 & 12.93 \\
\rowcolor{black!3}
InternVL3-38B & 32.27 & 58.51 & 9.22 & 26.15 & 60.51 & 13.33 \\
\bottomrule
\end{tabularx}

\caption{\textbf{Transcription outcomes on clear and moderately blurred text.} All rates are percentages. The 15 aliases use full images and the neutral prompt and are ordered by clear RR within each group.}
\label{tab:main781}

\endgroup
\end{table}

\paragraph{\textbf{Every alias returns the conventional word on clear print.}}
Table~\ref{tab:main781} reports the full panel on clear images.
RR runs from 8.45\% for Gemini 3.1 Flash Lite to 58.51\% for InternVL3-38B, with a 15-alias mean of 26.35\%.
Giving each model family equal weight yields a similar clear-image RR of 27.16\%.
These rates describe the constructed stress set under the recorded provider configurations, not general OCR accuracy.

\paragraph{\textbf{Closed-source models show lower rewriting overall.}}
On clear images, the six closed-source models have a mean RR of 14.94\%, compared with 33.96\% for the nine open-source models. Their mean literal accuracy is also higher, at 83.12\% versus 57.19\%. Under moderate blur, the same pattern remains: mean RR is 27.89\% for closed-source models and 42.96\% for open-source models. These differences are descriptive, since the two groups differ in model family, scale, and training setup.

\paragraph{\textbf{Rewriting rate does not restate literal accuracy.}}
Models that rewrite equally often can read very different amounts of text correctly.
GPT-5.5 and Qwen3-VL-8B Instruct both rewrite 19.46\% of targets, yet their literal accuracies are 78.10\% and 62.23\%, with O rates of 2.43\% and 18.31\%.
The inversion also runs the other way.
GPT-5.2 rewrites more than Qwen3-VL-32B Instruct (29.58\% versus 27.53\%) while reading more targets correctly (67.35\% versus 60.95\%). GLM-4.6V rewrites more than Qwen3-VL-8B Instruct (21.25\% versus 19.46\%) at 12 points higher literal accuracy.
Ranking on RR alone would therefore order these pairs wrongly for either purpose.
Table~\ref{tab:main781} reports all three outcomes for both conditions, and Appendices~\ref{app:main-statistics} and~\ref{app:supporting-data} give uncertainty estimates and supporting plots.

\paragraph{\textbf{Thinking aliases are associated with higher rewriting.}}
\label{sec:thinking}

Within Qwen3-VL, same image comparisons under identical prompts and recorded decoding settings give higher RR for Thinking than Instruct at the two smaller sizes. 
The differences are 18.44 and 16.90 percentage points, and both survive Holm correction across three sizes.
Literal accuracy falls by a similar amount, from 62.23\% to 48.53\% at 8B and from 60.95\% to 47.50\% at 32B.
At 235B-A22B the RR difference is $-0.13$ points (95\% CI $[-2.94,2.69]$), where 49 literal-to-canonical changes are offset by 54 in the reverse direction.
Mixture-of-experts architecture might be a reason why the results were not noticeable.
Archived reasoning text shows what the gap looks like in individual cases.
For ``uher'', the 8B Thinking trace repeats the printed candidate, questions it, and answers ``usher'', while Instruct returns ``uher''.
All 2,343 Thinking outputs retain reasoning text, and seven canonical-answer traces contain both exact candidates.
These traces show behavioral differences but do not explain their internal cause.

\paragraph{\textbf{What does \emph{Other} contain?}}
We reviewed 100 of the 713 \emph{Other} outputs, sampled proportionally across 15 models. Most were near-target spelling errors (66\%); 30\% included extra text and 4\% named a different concept. 

\subsection{Context Effect Analysis}
\label{sec:peripheral-input}

The preceding results show that rewriting occurs, but they do not isolate the effect of the surrounding scene because the target text and context appear together.
We analyze the effect of scene context through the scene-removal experiment on SceneFaith and the evaluation on the Controlled Fixed-Target-Patch Set.

\paragraph{\textbf{The effect of removing the surrounding scene.}}
\label{sec:gray-mask}

Gray masking replaces all pixel outside the red-box rectangle with RGB $(128,128,128)$ while keeping the target region, size, and position unchanged.
Masking removes all surrounding information at once, including semantic cues, nearby text, and clutter. 
Pixel replay verifies all 781 inputs, and the four models provide 3,124 accepted responses under fixed decoding settings.
RR falls by 18.57, 6.53, 9.60, and 20.10 percentage points for GPT, Gemini, Claude, and Qwen, and all paired 95\% intervals exclude zero (Table~\ref{tab:context-mask781}).
Literal accuracy rises in parallel, by 16.90, 7.17, 11.78, and 20.36 points.
This increase reflects recovery rather than a shift to other errors: many canonical outputs become literal after masking, few become other errors. 
Correct transcription after changing only the background shows that the model's output is influenced by the surrounding scene.

\begin{table}[t]
\centering\begingroup
\footnotesize\fontencoding{OT1}\fontfamily{ptm}\selectfont

\vspace{3pt}
\renewcommand{\arraystretch}{1.14}
\setlength{\tabcolsep}{2.7pt}
\begin{tabularx}{\linewidth}{>{\raggedright\arraybackslash}Xrrrrrrrc}
\toprule
\multirow{2}{*}{\textbf{Model alias}} & \multicolumn{3}{c}{\textbf{Full image}} & \multicolumn{3}{c}{\textbf{Gray mask}} & \multirow{2}{*}{$\Delta$C} & \multirow{2}{*}{95\% CI} \\
\cmidrule(lr){2-4}\cmidrule(lr){5-7}
 & L & C & O & L & C & O & & \\
\midrule
GPT-5.2 & 67.35 & 29.58 & 3.07 & 84.25 & 11.01 & 4.74 & -18.57 & [-21.64, -15.49] \\
Gemini 3.5 Flash & 89.76 & 8.96 & 1.28 & 96.93 & 2.43 & 0.64 & -6.53 & [-8.45, -4.74] \\
Claude Sonnet 4.6 & 85.02 & 12.16 & 2.82 & 96.80 & 2.56 & 0.64 & -9.60 & [-11.91, -7.43] \\
Qwen3-VL-32B T. & 47.50 & 44.43 & 8.07 & 67.86 & 24.33 & 7.81 & -20.10 & [-23.30, -17.03] \\
\bottomrule
\end{tabularx}
\caption{\textbf{Scene removal with fixed target pixels.} Intervals use 20,000 paired-image bootstrap resamples.}
\label{tab:context-mask781}
\endgroup
\end{table}


\paragraph{\textbf{Cropping reduces rewriting and recovers many difficult cases.}}
\label{sec:crop}

\begin{table}[t]
\centering
\begingroup
\small\fontencoding{OT1}\fontfamily{ptm}\selectfont

\vspace{3pt}
\renewcommand{\arraystretch}{1.14}
\setlength{\tabcolsep}{4pt}
\begin{tabularx}{\linewidth}{>{\raggedright\arraybackslash}Xrrrrrr}
\toprule
\multirow{2}{*}{\textbf{Model alias}} & \multicolumn{3}{c}{\textbf{Full image}} & \multicolumn{3}{c}{\textbf{Crop only}} \\
\cmidrule(lr){2-4}\cmidrule(l){5-7}
 & L (Acc.) $\uparrow$ & C (RR) $\downarrow$ & O $\downarrow$ & L (Acc.) $\uparrow$ & C (RR) $\downarrow$ & O $\downarrow$ \\
\midrule
\rowcolor{black!3}
GPT-5.2 & 67.35 & 29.58 & 3.07 & 89.88 & 5.63 & 4.48 \\
Gemini 3.5 Flash & 89.76 & 8.96 & 1.28 & 96.93 & 2.18 & 0.90 \\
\rowcolor{black!3}
Claude Sonnet 4.6 & 85.02 & 12.16 & 2.82 & 93.60 & 3.59 & 2.82 \\
Qwen3-VL-32B Thinking & 47.50 & 44.43 & 8.07 & 86.94 & 7.55 & 5.51 \\
\bottomrule
\end{tabularx}
\caption{\textbf{Transcription outcomes for full-image and crop-only recognition.} All rates are percentages, with 781 images per entry.}
\label{tab:crop781}
\endgroup
\end{table}

Crop-only recognition reduces RR to 2.18--7.55\% and improves literal accuracy for all four models (Table~\ref{tab:crop781}).
Each crop keeps the red outline and a small margin around the target, and is upsampled when needed.
This produces 3,124 valid responses under the same request settings.
Cropping changes not only the surrounding scene, but also the framing and target scale.
Therefore, the lower RR reflects sensitivity to the input view rather than to scene content alone.

\paragraph{\textbf{Controlled Fixed-Target-Patch Set Analysis.}}
\label{sec:pixel150}

In this set, we test sensitivity to surrounding context in two ways while keeping the target pixels unchanged: retaining the original background or replacing it with a scene semantically unrelated to the target word. 
The 150-pair evaluation varies the surrounding without removing it, holding the target patch and its scale identical across the two scenes.
Mean RR rises from 6.85\% in identifier scenes to 11.15\% in semantic scenes, while literal accuracy falls from 89.70\% to 85.39\% and mean O stays at 3.45\%.
Nine of eleven aliases show positive differences, and five have 95\% paired-bootstrap intervals above zero before multiplicity correction.
GPT-5.2 shows the largest effect, at $+10.00$ points.
The result is consistent with the masking experiment, even though both conditions contain surrounding content.
All evaluated aliases show increased rewriting in matching scenes compared with shuffled scenes, suggesting a consistent effect of the surrounding scene. 
However, the semantic and visual sources of this effect remain entangled.

\subsection{Target Blur Increases Rewriting}
\label{sec:blur}

Section~\ref{sec:peripheral-input} keep the target text unchanged. 
In contrast, target-local blur weakens the target text while keeping the surrounding scene and red outline unchanged.
For box height $h$, Gaussian standard deviations in clear image pixels are
\[
\sigma_{\mathrm{mild}}=\max(0.6,0.015h),\quad
\sigma_{\mathrm{moderate}}=\max(1.2,0.030h),\quad
\sigma_{\mathrm{strong}}=\max(2.4,0.060h).
\]
We conduct a moderate-blur evaluation on all 15 models and a three-level blur evaluation on four selected models.
All 15 aliases rewrite more under moderate blur, and rewriting rises monotonically with severity in the four-model experiment.
Every alias in Table~\ref{tab:main781} has a higher RR under moderate blur than on clear images.
The RR increase remains similar with equal family weighting ($+11.15$ points) and after removing Qwen ($+11.60$ points). Literal accuracy also drops by 13.76 points (Appendix~\ref{app:family-sensitivity}). 
This suggests that the pattern is not driven by the model panel composition. 
In the four-model experiment, RR increases from moderate to strong blur for all models, reaching 43.02--60.69\% under strong blur (Table~\ref{tab:blur-severity781}). 

\subsection{Analyzing Lexical Priors in Rewriting}
\label{sec:string-properties}

The interventions above vary the input for a fixed set of targets. 
Rewriting also varies across targets under identical conditions, and the following analyses relate that variation to external lexical preference, string length, and the specific character edit.
All three use clear image outputs and are exploratory associations on a constructed set, not causal claims.
\paragraph{\textbf{Stronger lexical preference is associated with more rewriting.}}
\label{sec:lexical-prior}
We use DistilGPT2 \citep{sanh2020distilbert} to estimate semantic-prior strength $A_i=\ell(c_i)$ and the preference gap $D_i=\ell(c_i)-\ell(y_i)$, 
\begin{equation}
\ell(w)=\frac{1}{K}
\sum_{t=1}^{K}\log P_{\mathrm{LM}}\bigl(b(w)\mid T_t\bigr),
\qquad
A_i=\ell(c_i),\quad D_i=\ell(c_i)-\ell(y_i).
\label{eq:lexical-prior}
\end{equation}
using average log-likelihood over four neutral prefixes without access to images or OCR outputs.
Across 781 images and 15 aliases, $D$ is positively correlated with mean RR ($\rho=0.255$). 
When the samples are divided into five groups by $D$, RR is 20.47\% in the lowest group and 38.03\% in the highest group.
The association remains after controlling for other factors, including model, domain, string properties, and box geometry.
The preference gap $D$ matters more than the absolute score $A$ of the conventional string, whose association is unclear.
These use external lexical proxies rather than any VLM's internal prior; Appendix~\ref{app:lexical-prior} reports detailed results and additional sensitivity checks.

\paragraph{\textbf{Longer conventional strings tend to be rewritten.}}
RR rises from 14.27\% for conventional strings of 4--5 characters to 45.02\% for strings of at least 10, while literal accuracy falls from 79.80\% to 49.79\%.
The positive association appears in all 15 models. 
Each additional character is associated with a 23\% increase in the odds of rewriting.
The association remains after controlling for lexical scores or perturbation type, suggesting that the length effect is not explained by either factor.
Appendix~\ref{app:string-length} specifies controls and full L/C/O results.

\paragraph{\textbf{Lowercase \texttt{l}/\texttt{i} substitutions are rewritten at 59.78\%, far above other letter-shape substitutions.}}
\label{sec:character-confusions}
Among letter-shape substitutions, RR is 59.78\% for the 31 lowercase \texttt{l}$\leftrightarrow$\texttt{i} images versus 37.17\% for the other 120.
These two characters are visually similar and lexically substitutable, so the result is compatible with both perceptual misreading and lexical correction.
(Appendix~\ref{app:noise-types}).

\subsection{Post-Training Mitigation Pilot}
\label{sec:post-training-pilot}
We compare SFT and SFT+GRPO using Qwen3-VL-4B-Instruct trained on 300 clear--blurred image pairs (600 images).
Both methods use task specific prompts and are evaluated
on 781 test pairs (1,562 images) with greedy decoding.
The reward function is defined below:
\begin{equation}
z(\hat{y})=
\left[
\lambda\mathbf[{\hat{y}=y^\star}]
+
(1-\lambda)\max\left(
0,1-\frac{d_{\mathrm{edit}}(\hat{y},y^\star)}{|y^\star|}
\right)
\right]
\end{equation}
where $\hat{y}$ is the prediction, $y^\star$ is the target,
and $d_{\mathrm{edit}}$ denotes Levenshtein distance~\citep{lcvenshtcin1966binary}.
We remove only leading and trailing whitespace before comparison and set $\lambda=0.8$ for the exact-match term. 
The reward function $z(\hat{y})$ allows incorrect responses to receive partial credit based on their similarity to the target.
The exact-match term rewards correct outputs, while the similarity term provides partial credit for incorrect outputs.
GRPO uses the frozen SFT policy as its reference, with a KL coefficient of $0.04$.
\begin{table*}[t]
\centering
\small

\begin{tabular}{lccccc}
\toprule
Method & Overall EM $\uparrow$ & Clear EM $\uparrow$
& Blurred EM $\uparrow$ & Pair accuracy $\uparrow$ & CER $\downarrow$ \\
\midrule
Base Model
& $67.93$ & $66.07$ & $69.78$ & $42.25$ & $7.77$ \\
SFT
& $81.90 \pm 0.50$ & $83.35 \pm 0.68$
& $80.45 \pm 1.60$ & $66.58 \pm 0.59$ & $5.80 \pm 0.53$ \\
SFT+GRPO
& $\mathbf{82.44} \pm 0.33$ & $\mathbf{84.08} \pm 0.64$ 
& $\mathbf{80.79} \pm 1.22$ & $\mathbf{67.65} \pm 0.77$ 
& $\mathbf{5.64} \pm 0.34$ \\
\bottomrule
\end{tabular}
\caption{Results with 300 training pairs.
Values are percentages, reported as mean $\pm$ sample
standard deviation over three training seeds.
Pair accuracy requires both predictions in a pair to be correct.
}
\label{tab:sft_grpo_300pairs}
\end{table*}
As shown in Table~\ref{tab:sft_grpo_300pairs},
SFT+GRPO improves overall exact match (EM) by $0.53$
percentage points and pair accuracy by $1.07$ percentage
points, based on unrounded results.
Character error rate (CER) is the total character edit distance divided by the total target length; lower values indicate better performance. CER decreases from 5.80\% with SFT to 5.64\% with SFT+GRPO.
Overall EM improves across all three seeds, showing a modest
but consistent gain over SFT.
\section{Discussion}
\label{sec:diagnostic-discussion}

\paragraph{Surrounding context affects transcription fidelity.}
SceneFaith shows that transcription depends on more than the target characters.
Gray masking improves literal accuracy while keeping the target pixels unchanged, and the fixed-patch experiment shows that changing the surrounding scene can also change the output.
Cropping and lexical analyses further show that transcription is sensitive to image presentation, and lexical preference.
The different behaviors of the Qwen Thinking variants also suggest that character fidelity should be studied within the same model family.

\paragraph{Models could balance character evidence and context.}
A reliable recognizer should preserve clear characters while using context only when the text is ambiguous.
Crop recovery shows that some clear-image errors remain recoverable under a different view.
Under blur, rewriting still increases after the surrounding scene is removed in three of four models, suggesting that both lexical completion and visual ambiguity may contribute.
Future evaluations should therefore compare helpful, neutral, and misleading contexts around the same degraded target.

\paragraph{Rewriting and recovery should be measured together.}
A lower RR does not always mean better transcription: canonical outputs may become either correct literal outputs ($C\!\to\!L$) or other errors ($C\!\to\!O$).
Reporting L/C/O together separates these cases and keeps literal accuracy as the main measure of recovery. Section~\ref{sec:post-training-pilot} reports a pilot study on mitigating rewriting and improving transcription faithfulness.

\section{Limitations}

SceneFaith focuses on generated scenes with deliberately perturbed text;
generalization to natural images and broader OCR tasks remains to be tested. 
Scene interventions combine semantic and visual changes, while external lexical scores provide indirect evidence about internal mechanisms. 
Blurred targets lack complete readability validation, making information loss difficult to separate from lexical completion. 
Separate API runs may also introduce provider variation. 
The post-training pilot uses one model;
further evaluation requires matched compute, independent test data, and additional models.
\section{Conclusion}

SceneFaith makes canonical substitution auditable through 781 images and Literal/Canonical/Other outcomes across 15 model aliases.
Fixed-target scene removal reduces rewriting and improves literal accuracy in four models,
while complementary diagnostics characterize sensitivity to scene substitutions, image views, degradation, and lexical preference.
These findings establish peripheral-input sensitivity as a practical OCR reliability concern in the evaluated conditions.
Evidence-calibrated recognition requires preserving clear text and recovering the printed string from ambiguous inputs with reliable context.
SceneFaith provides a basis for testing transcription fidelity.
Future work can extend it to test whether reliable context helps recover degraded text.

\subsection*{AI Use Statement}
Generative models assisted with benchmark construction, experiment design, code and paper polishing.
We checked the AI work and take responsibility for the final content.

\subsection*{Ethics Statement}
The study uses generated scenes and model outputs, with no personal data or human-subject annotations.
Replacing visible text with familiar spellings may cause exact-match errors; retained raw responses make them traceable.

\subsection*{Reproducibility Statement}
Retained manifests, image hashes, prompts, parser, raw responses, and deterministic audits support the reported results.
Subsequent appendices document the external scorer, paired-image statistics, fixed-patch, consensus, and blur/mask audits.
The 781-image analyses use 15 aliases; the separate 150-pair diagnostic uses 11.

\bibliographystyle{iclr2027_conference}
\bibliography{iclr2027_conference}

\clearpage
\appendix

\section{Construction Inventory and Audit}
\label{app:construction}

The 781 PNGs retain their native resolution, unique IDs, printed and canonical strings, and file and RGB hashes. 
The printed string is the transcription gold; the canonical string is used only for analysis. 
Model-assisted screening, blind target readings, and visual review excluded unclear or missing targets, unsuitable scenes, and answer leakage. 
Hash audits verified all 781 crops and 2,343 blur transforms. These checks confirm pixel integrity, not readability.

\begin{table}[h]
\centering\small
\begin{tabular}{@{}lrlrlr@{}}
\toprule
Category  & Images & Category  & Images & Category  & Images \\
\midrule
Animal & 109 & Electronics & 14 & Profession & 79 \\
Building & 20 & Home & 27 & Recipe & 91 \\
City & 64 & Landform & 5 & Sport & 44 \\
Clothing & 26 & Medicine & 37 & Tool & 14 \\
Command & 32 & Music & 42 & Vehicle & 34 \\
Country & 56 & Plant & 87 & Total & 781 \\
\bottomrule
\end{tabular}
\caption{Category inventory of the complete 781-image benchmark. All reported experiments use this aggregate collection.}
\label{tab:domains781}
\end{table}


\section{Lexical Prior Analysis}
\label{app:lexical-prior}

We examine associations between external language-model scores and 11,715 clear-image outputs from 15 models and 781 images. The score definition was fixed before linking scores to OCR outcomes; the analyses are exploratory.

\subsection{Scoring and Estimation}

Let $y_i$ be the printed string and $c_i$ its conventional spelling. We use DistilGPT2 to score each string after four fixed neutral prefixes:
\begin{equation}
\ell(w)=\frac{1}{K}
\sum_{t=1}^{K}\log P_{\mathrm{LM}}\bigl(b(w)\mid T_t\bigr),
\qquad
A_i=\ell(c_i),\quad D_i=\ell(c_i)-\ell(y_i).
\label{eq:lexical-prior}
\end{equation}
where $b(w)$ adds a leading space and a final period. Thus, $A_i$ measures the conventional string's likelihood, while $D_i$ measures its advantage over the printed string.

Correlations and quintiles use each image's mean rewriting rate across the 15 models. 
Regressions use individual outputs and adjust for model, domain, string length, edit distance, nonalphabetic characters, and target-box geometry. 
Intervals account for repeated outputs from the same image.

\subsection{Associations and Sensitivity}

The preference gap $D$ correlates with rewriting ($\rho=0.255$, 95\% CI $[0.189,0.321]$). Rewriting rises from 20.47\% in the lowest quintile to 38.03\% in the highest, a difference of 17.57 points ($[11.86,23.23]$). In the joint adjusted model, $D$ remains associated with rewriting, whereas $A$ does not (Tables~\ref{tab:lexical781} and~\ref{tab:distilgpt2-two-priors}). The association for $D$ persists across the reported scoring and string-subset checks.
These associations involve external scores and do not identify a VLM-internal lexical mechanism.

\begin{table}[t]\centering\small
\begin{tabular}{@{}lrrr@{}}\toprule
Analysis & Words & Adjusted OR & 95\% CI \\\midrule
Primary: period-terminated gap & 781 & 1.352 & [1.197, 1.527] \\
No-period gap & 781 & 1.377 & [1.214, 1.562] \\
No-period gap, per character & 781 & 1.341 & [1.195, 1.506] \\
Period gap, per character & 781 & 1.312 & [1.169, 1.473] \\
Equal-length candidates & 398 & 1.387 & [1.190, 1.617] \\
Alphabetic printed strings & 704 & 1.340 & [1.183, 1.519] \\
Template 1 only & 781 & 1.341 & [1.192, 1.509] \\
Template 2 only & 781 & 1.321 & [1.171, 1.490] \\
Template 3 only & 781 & 1.303 & [1.159, 1.465] \\
Template 4 only & 781 & 1.348 & [1.195, 1.520] \\
\bottomrule\end{tabular}
\caption{Exploratory lexical-preference analysis and 9 sensitivity checks.}
\label{tab:lexical781}\end{table}

\begin{table}[t]
\centering\small
\begin{tabular}{@{}lrcc@{}}
\toprule
Analysis & Words & Strength $A$: OR [95\% CI] & Gap $D$: OR [95\% CI] \\
\midrule
All: separate predictors & 781 & 1.202 [1.081, 1.337] & 1.352 [1.197, 1.527] \\
All: joint predictors & 781 & 1.017 [0.895, 1.157] & 1.339 [1.155, 1.551] \\
Alphabetic only & 704 & 1.225 [1.099, 1.364] & 1.340 [1.183, 1.519] \\
Equal length & 398 & 1.179 [1.008, 1.379] & 1.387 [1.190, 1.617] \\
\bottomrule\end{tabular}
\caption{DistilGPT2 strength $A$ and preference gap $D$ on the 781-image benchmark.}
\label{tab:distilgpt2-two-priors}\end{table}

\subsection{String Length}
\label{app:string-length}

Rewriting rises from 14.27\% for conventional strings of 4--5 characters to 45.02\% for strings of at least 10 characters (Table~\ref{tab:length781}). The adjusted odds ratio is 1.230 per additional character ($[1.157,1.309]$), or 1.127 ($[1.039,1.223]$) after adding $A$ and $D$.

\begin{table}[htbp]\centering\small
\begin{tabular}{@{}lrrrrr@{}}\toprule
Characters in $c_i$ & Images & L (\%) & C / RR (\%) & O (\%) & RR 95\% CI \\\midrule
4--5 & 199 & 79.80 & 14.27 & 5.93 & [11.99, 16.68] \\
6--7 & 302 & 69.16 & 24.48 & 6.36 & [21.88, 27.17] \\
8--9 & 199 & 60.13 & 33.67 & 6.20 & [30.05, 37.39] \\
$\geq 10$ & 81 & 49.79 & 45.02 & 5.19 & [39.26, 50.95] \\
\bottomrule\end{tabular}
\caption{Longer conventional strings accompany more canonical substitutions.}
\label{tab:length781}\end{table}

\section{Supplementary Statistics for the Clear-Image Evaluation}
\label{app:main-statistics}

Table~\ref{tab:main781-statistics} reports rewriting rates (RR) for 15 aliases on 781 clear images.
Domain-macro RR weights 17 categories equally; the model mean weights aliases equally.
The 95\% confidence interval reflects uncertainty due to the limited number of images.
Appendix~\ref{app:family-sensitivity} tests family weighting.

\begin{table}[htbp]
\centering
\begingroup
\small\fontencoding{OT1}\fontfamily{ptm}\selectfont

\vspace{3pt}
\renewcommand{\arraystretch}{1.14}
\setlength{\tabcolsep}{5pt}
\begin{tabularx}{\linewidth}{>{\raggedright\arraybackslash}Xrrr}
\toprule
\multirow{2}{*}{\textbf{Model alias}} & \multicolumn{3}{c}{\textbf{Rewriting rate (\%)}} \\
\cmidrule(l){2-4}
 & Overall $\downarrow$ & 95\% Wilson CI & Category macro $\downarrow$ \\
\midrule
\rowcolor{black!3}
Gemini 3.1 Flash & 8.45 & \textcolor{black!65}{[6.70, 10.61]} & 8.33 \\
Gemini 3.5 Flash & 8.96 & \textcolor{black!65}{[7.16, 11.17]} & 7.43 \\
\rowcolor{black!3}
Gemini 3 Flash & 11.01 & \textcolor{black!65}{[9.00, 13.40]} & 9.34 \\
Claude Sonnet 4.6 & 12.16 & \textcolor{black!65}{[10.05, 14.64]} & 11.27 \\
\rowcolor{black!3}
GPT-5.5 & 19.46 & \textcolor{black!65}{[16.84, 22.39]} & 18.33 \\
Qwen3-VL-8B Instruct & 19.46 & \textcolor{black!65}{[16.84, 22.39]} & 19.06 \\
\rowcolor{black!3}
GLM-4.6V & 21.25 & \textcolor{black!65}{[18.53, 24.26]} & 19.64 \\
Qwen3-VL-32B Instruct & 27.53 & \textcolor{black!65}{[24.51, 30.77]} & 26.90 \\
\rowcolor{black!3}
GPT-5.2 & 29.58 & \textcolor{black!65}{[26.48, 32.87]} & 27.80 \\
Kimi K2.5 & 31.88 & \textcolor{black!65}{[28.71, 35.23]} & 29.51 \\
\rowcolor{black!3}
Qwen3-VL-235B-A22B Thinking & 32.27 & \textcolor{black!65}{[29.08, 35.62]} & 31.40 \\
Qwen3-VL-235B-A22B Instruct & 32.39 & \textcolor{black!65}{[29.21, 35.76]} & 34.16 \\
\rowcolor{black!3}
Qwen3-VL-8B Thinking & 37.90 & \textcolor{black!65}{[34.56, 41.35]} & 38.82 \\
Qwen3-VL-32B Thinking & 44.43 & \textcolor{black!65}{[40.98, 47.93]} & 42.17 \\
\rowcolor{black!3}
InternVL3-38B & 58.51 & \textcolor{black!65}{[55.03, 61.92]} & 58.36 \\
\midrule
\rowcolor{black!8}
\textbf{Model mean} & \textbf{26.35} & -- & \textbf{25.50} \\
\bottomrule
\end{tabularx}
\caption{Supplementary statistics for clear-image rewriting. }
\label{tab:main781-statistics}
\endgroup
\end{table}

\section{Sensitivity to Model-Family Composition}
\label{app:family-sensitivity}

We average aliases within each of seven model families, then weight families equally.
Qwen has six aliases, Gemini three, GPT two, and Claude, GLM, InternVL, and Kimi one each.
All comparisons use the same 781 images with accepted clear and moderate-blur outputs for every alias.
Pointwise 95\% intervals use 10,000 paired-image bootstrap resamples; aliases and families stay fixed.

\begin{table}[htbp]
\centering\begingroup
\small\fontencoding{OT1}\fontfamily{ptm}\selectfont

\vspace{3pt}
\renewcommand{\arraystretch}{1.14}
\setlength{\tabcolsep}{3pt}
\begin{tabularx}{\linewidth}{>{\raggedright\arraybackslash}Xrrrrrr}
\toprule
\textbf{Weighting} & \multicolumn{3}{c}{\textbf{Clear (\%)}} & \multicolumn{3}{c}{\textbf{Moderate (\%)}} \\
\cmidrule(lr){2-4}\cmidrule(l){5-7}
 & L & C & O & L & C & O \\
\midrule
Equal alias weighting (15 aliases) & 67.55 & 26.36 & 6.09 & 54.94 & 36.95 & 8.11 \\
Equal family weighting (7 families) & 68.03 & 27.16 & 4.81 & 54.65 & 38.31 & 7.05 \\
Excluding Qwen: equal alias weighting (9 aliases) & 74.52 & 22.36 & 3.12 & 60.95 & 34.40 & 4.64 \\
Excluding Qwen: equal family weighting (6 families) & 69.86 & 26.29 & 3.85 & 56.10 & 37.90 & 6.00 \\
\bottomrule
\end{tabularx}
\vspace{7pt}
\begin{tabularx}{\linewidth}{>{\raggedright\arraybackslash}Xccc}
\toprule
\textbf{Paired change (pp)} & $\Delta$L & $\Delta$C & $\Delta$O \\
\midrule
Equal alias weighting (15 aliases) & \shortstack{-12.61\\{\scriptsize [-13.82, -11.44]}} & \shortstack{+10.59\\{\scriptsize [+9.44, +11.74]}} & \shortstack{+2.02\\{\scriptsize [+1.14, +2.94]}} \\
Equal family weighting (7 families) & \shortstack{-13.39\\{\scriptsize [-14.72, -12.08]}} & \shortstack{+11.15\\{\scriptsize [+9.87, +12.43]}} & \shortstack{+2.24\\{\scriptsize [+1.32, +3.22]}} \\
Excluding Qwen: equal alias weighting (9 aliases) & \shortstack{-13.56\\{\scriptsize [-14.89, -12.25]}} & \shortstack{+12.04\\{\scriptsize [+10.81, +13.29]}} & \shortstack{+1.52\\{\scriptsize [+0.81, +2.28]}} \\
Excluding Qwen: equal family weighting (6 families)& \shortstack{-13.76\\{\scriptsize [-15.16, -12.36]}} & \shortstack{+11.60\\{\scriptsize [+10.25, +12.96]}} & \shortstack{+2.15\\{\scriptsize [+1.24, +3.13]}} \\
\bottomrule
\end{tabularx}
\endgroup
\caption{Sensitivity to model-family weighting. Brackets show 95\% paired-image bootstrap confidence intervals for changes from clear to moderate blur.}
\label{tab:family-sensitivity781}
\end{table}

In all seven leave-one-family-out analyses, moderate blur raises C by 9.97--12.66 points and lowers L by 12.07--14.59.

\section{Qwen Instruct--Thinking Comparisons Analysis}
\label{app:qwen-thinking}

\paragraph{Paired comparisons.}
Each Instruct--Thinking pair covers the same 781 images at one of three Qwen sizes.
Pointwise 95\% intervals use 20,000 paired-image bootstrap resamples; exact McNemar tests compare aliases within each size, with Holm correction across sizes.
Thinking has lower literal accuracy at 8B and 32B, but no clear change at 235B-A22B.
The 235B-A22B model uses a mixture-of-experts (MoE) architecture.
The respective accuracy changes are $-13.70$ ($[-17.67,-9.86]$), $-13.44$ ($[-16.90,-9.86]$), and $+1.15$ ($[-1.79,4.10]$) points.
Scoring whole answers gives similar RR gaps: $+18.95$, $+17.67$, and $+0.38$ points.
These are comparisons between distinct served aliases, not a thinking toggle within one model.

\begin{table}[H]
\centering\small
\setlength{\tabcolsep}{4pt}
\begin{tabular}{@{}llrrr@{}}
\toprule
Size & Variant & Literal & Canonical & Other \\
\midrule
8B & Instruct & 486 (62.23) & 152 (19.46) & 143 (18.31) \\
8B & Thinking & 379 (48.53) & 296 (37.90) & 106 (13.57) \\
32B & Instruct & 476 (60.95) & 215 (27.53) & 90 (11.52) \\
32B & Thinking & 371 (47.50) & 347 (44.43) & 63 (8.07) \\
235B-A22B & Instruct & 478 (61.20) & 253 (32.39) & 50 (6.40) \\
235B-A22B & Thinking & 487 (62.36) & 252 (32.27) & 42 (5.38) \\
\bottomrule
\end{tabular}
\caption{Qwen clear-image outcome counts (percentages), with $N=781$ for every row.}
\label{tab:qwen-thinking-lco781}
\end{table}

\begin{table}[H]
\centering\small
\setlength{\tabcolsep}{4pt}
\begin{tabular}{@{}llrrr@{}}
\toprule
Size & Instruct outcome & Thinking L & Thinking C & Thinking O \\
\midrule
8B & L & 304 & 136 & 46 \\
8B & C & 16 & 128 & 8 \\
8B & O & 59 & 32 & 52 \\
32B & L & 317 & 134 & 25 \\
32B & C & 23 & 183 & 9 \\
32B & O & 31 & 30 & 29 \\
235B-A22B & L & 412 & 49 & 17 \\
235B-A22B & C & 54 & 190 & 9 \\
235B-A22B & O & 21 & 13 & 16 \\
\bottomrule
\end{tabular}
\caption{Matched-image outcome counts from Instruct (row) to Thinking (column). }
\label{tab:qwen-thinking-transitions781}
\end{table}

\paragraph{Reasoning text.}
All 2,343 Thinking outputs retain reasoning text; Instruct outputs do not.
Exact searches find both the printed and canonical strings in seven canonical-answer traces (1, 1, and 5 by size).
An 8B trace changes ``uher'' to ``usher''; a 32B trace identifies ``mont-evideo'' as a likely typo but copies it.
These traces suggest that the model may sometimes recognize the printed text but still produce an incorrect final answer.

\paragraph{Exploratory checks.}
Moderate blur changes the Thinking--Instruct RR gap by $+2.18$ ($[-1.54,6.02]$), $+0.77$ ($[-2.94,4.48]$), and $+1.66$ ($[-1.79,5.12]$) points by size.
The intervals leave blur amplification unresolved.
Regressions against external lexical gap $D$ show no positive association at 8B or 32B; the 235B slope is $-3.56$ points per SD ($[-6.29,-0.94]$).
These exploratory intervals are unadjusted for multiple checks, and the associations do not identify a lexical mechanism.

\section{Character Perturbations and Canonical Rewriting}
\label{app:noise-types}

We classified the printed-to-conventional edits in all 781 images into eight groups, independently of model outputs (Table~\ref{tab:noise-types781}).
Fixed mappings define letter-shape substitutions (\texttt{l/i}, \texttt{e/c}, \texttt{u/v}, \texttt{n/m}, \texttt{h/b}, \texttt{a/o}, in both directions), digit/symbol substitutions (\texttt{o,i,l,s,a,g,b} to \texttt{0,1,1,5,@,9,6}), and split/merge edits (\texttt{m/rn}, \texttt{w/vv}, \texttt{cl/d}).
The other groups are duplication, deletion, insertion, transposition, and other substitution; inserting an adjacent identical character counts as duplication.

\begin{table}[htbp]
\centering\small

\setlength{\tabcolsep}{4pt}
\renewcommand{\arraystretch}{1.12}
\begin{tabularx}{\linewidth}{@{}Xrrrrr@{}}
\toprule
Perturbation & $n$ & Acc. & RR & Other & RR 95\% CI \\
\midrule
Letter-shape substitution & 151 & 53.02 & 41.81 & 5.17 & [37.31, 46.23] \\
Character split / merge & 26 & 50.26 & 38.72 & 11.03 & [28.72, 48.72] \\
Character duplication & 149 & 61.21 & 33.47 & 5.32 & [29.71, 37.23] \\
Digit / symbol substitution & 61 & 66.67 & 27.98 & 5.36 & [20.98, 35.30] \\
Character deletion & 172 & 77.09 & 18.18 & 4.73 & [15.23, 21.28] \\
Character insertion & 36 & 77.04 & 17.59 & 5.37 & [10.56, 25.56] \\
Adjacent transposition & 141 & 75.37 & 15.93 & 8.70 & [13.14, 18.82] \\
Other letter substitution & 45 & 80.15 & 12.44 & 7.41 & [8.15, 17.19] \\
\midrule
\multicolumn{6}{l}{\textit{Selected character substitutions: conventional $\to$ printed}} \\
\texttt{l}$\to$\texttt{i} & 10 & 37.33 & 60.00 & 2.67 & [40.00, 79.33] \\
\texttt{i}$\to$\texttt{l} & 21 & 36.51 & 59.68 & 3.81 & [50.48, 68.58] \\
\texttt{l}$\to$\texttt{1} & 7 & 41.90 & 56.19 & 1.90 & [36.19, 76.21] \\
\texttt{i}$\to$\texttt{1} & 16 & 61.67 & 27.50 & 10.83 & [15.42, 41.67] \\
\bottomrule
\end{tabularx}
\caption{Clear-image outcomes by recorded character perturbation.}
\label{tab:noise-types781}
\end{table}

For 31 lowercase \texttt{l}$\leftrightarrow$\texttt{i} images, RR exceeds that of the other 120 letter-shape images by 22.62 points ($[12.59,32.90]$).
The post hoc logistic model adjusts for model, domain, conventional length, box geometry, and external lexical scores $A$ and $D$ (Appendix~\ref{app:lexical-prior}).
Its image-clustered interval gives OR 3.39 ($[2.16,5.30]$).

\section{Blur Statistics}
\label{app:supporting-data}

Tables~\ref{tab:blur-severity781} and table~\ref{tab:moderate-effects781} report rewriting rates across blur levels and paired changes under moderate blur.
Pointwise 95\% intervals use 10,000 paired-image bootstrap resamples and describe image variation in the saved calls.

\begin{table}[htbp]
\centering
\begingroup
\small\fontencoding{OT1}\fontfamily{ptm}\selectfont

\vspace{3pt}
\renewcommand{\arraystretch}{1.14}
\setlength{\tabcolsep}{5pt}
\begin{tabularx}{\linewidth}{>{\raggedright\arraybackslash}Xrrrr}
\toprule
\textbf{Model alias} & Clear & Mild & Moderate & Strong \\
\midrule
GPT-5.2 & 29.58 & 30.09 & 45.33 & 60.69 \\
Gemini 3.5 Flash & 8.96 & 10.50 & 18.44 & 54.67 \\
Claude Sonnet 4.6 & 12.16 & 10.37 & 22.66 & 43.02 \\
Qwen3-VL-32B Thinking & 44.43 & 47.25 & 51.47 & 59.72 \\
\bottomrule
\end{tabularx}
\endgroup
\caption{Rewriting rates (\%) across blur severity. }
\label{tab:blur-severity781}
\end{table}

\begin{table}[H]
\centering\begingroup
\small\fontencoding{OT1}\fontfamily{ptm}\selectfont

\vspace{3pt}
\renewcommand{\arraystretch}{1.13}
\setlength{\tabcolsep}{3.5pt}
\begin{tabularx}{\linewidth}{>{\raggedright\arraybackslash}Xrrrc}
\toprule
\textbf{Model alias} & Clear & Moderate & $\Delta$RR & 95\% CI \\
\midrule
Gemini 3.1 Flash Lite & 8.45 & 20.74 & +12.29 & [+9.99, +14.72] \\
Gemini 3.5 Flash & 8.96 & 18.44 & +9.48 & [+7.04, +11.91] \\
Gemini 3 Flash Preview & 11.01 & 19.97 & +8.96 & [+6.53, +11.40] \\
Claude Sonnet 4.6 & 12.16 & 22.66 & +10.50 & [+7.43, +13.57] \\
GPT-5.5 & 19.46 & 40.20 & +20.74 & [+17.41, +24.07] \\
Qwen3-VL-8B I. & 19.46 & 26.76 & +7.30 & [+4.87, +9.86] \\
GLM-4.6V & 21.25 & 32.01 & +10.76 & [+8.32, +13.32] \\
Qwen3-VL-32B I. & 27.53 & 33.80 & +6.27 & [+3.59, +9.09] \\
GPT-5.2 & 29.58 & 45.33 & +15.75 & [+12.80, +18.82] \\
Kimi K2.5 & 31.88 & 49.68 & +17.80 & [+14.60, +21.13] \\
Qwen3-VL-235B T. & 32.27 & 43.28 & +11.01 & [+8.07, +14.08] \\
Qwen3-VL-235B I. & 32.39 & 41.74 & +9.35 & [+6.79, +12.04] \\
Qwen3-VL-8B T. & 37.90 & 47.38 & +9.48 & [+6.27, +12.68] \\
Qwen3-VL-32B T. & 44.43 & 51.47 & +7.04 & [+3.97, +10.24] \\
InternVL3-38B & 58.46 & 60.51 & +2.05 & [-0.38, +4.49] \\
\bottomrule
\end{tabularx}
\endgroup
\caption{Matched moderate-blur changes on the SceneFaith benchmark.}
\label{tab:moderate-effects781}
\end{table}

\section{Fixed-Target-Patch Diagnostic on 150 Pairs}
\label{app:pixel150}

We constructed 150 image pairs to test how surrounding scenes affect rewriting while keeping target pixels fixed.
Each pair places an identical target patch, including its local background, at the same position and scale on a $1024\times1024$ canvas.
Scene A supports the conventional word, while scene B presents the printed string as an identifier in a context that does not encourage rewriting.
The printed gold and prompts are identical within each pair.
All 11 models have complete responses for all 150 pairs.
RR differences are computed within matched pairs, with pointwise 95\% paired-image bootstrap intervals.

Across the 11 models, mean RR increases from 6.85\% in B to 11.15\% in A, a rise of 4.30 percentage points.
Nine models show positive differences, and five have intervals entirely above zero before adjustment for multiple comparisons (Table~\ref{tab:pixel150}).
These results show that rewriting responds to the surrounding scene even when the target pixels remain unchanged.

\begin{table}[htbp]
\centering
\small
\setlength{\tabcolsep}{4pt}

\begin{tabular}{lrrrrrr}
\toprule
Model alias & A: L & A: C & B: L & B: C & $\Delta$ & 95\% CI \\
\midrule
GPT-5.2 & 80.67 & 17.33 & 89.33 & 7.33 & $+10.00$ & $[4.00, 16.00]$ \\
GPT-5.5 & 80.67 & 16.00 & 90.00 & 8.00 & $+8.00$ & $[2.00, 14.67]$ \\
Claude Sonnet 4.6 & 99.33 & 0.67 & 98.00 & 0.67 & $+0.00$ & $[-2.00, 2.00]$ \\
Gemini 3.1 Flash Lite & 98.67 & 1.33 & 98.67 & 1.33 & $+0.00$ & $[-2.00, 2.00]$ \\
Gemini 3.5 Flash & 96.67 & 2.67 & 96.00 & 2.00 & $+0.67$ & $[-1.33, 3.33]$ \\
Qwen3-VL-8B-I & 88.00 & 8.00 & 92.67 & 4.00 & $+4.00$ & $[0.67, 8.00]$ \\
Qwen3-VL-8B-T & 76.67 & 15.33 & 79.33 & 12.00 & $+3.33$ & $[-1.33, 8.00]$ \\
Qwen3-VL-32B-I & 82.00 & 12.67 & 86.00 & 10.67 & $+2.00$ & $[-2.00, 6.02]$ \\
Qwen3-VL-32B-T & 71.33 & 22.00 & 77.33 & 15.33 & $+6.67$ & $[2.00, 12.00]$ \\
Qwen3-VL-235B-A22B-I & 79.33 & 16.67 & 90.00 & 7.33 & $+9.33$ & $[4.67, 14.67]$ \\
Qwen3-VL-235B-A22B-T & 86.00 & 10.00 & 89.33 & 6.67 & $+3.33$ & $[-0.67, 7.33]$ \\
\bottomrule
\end{tabular}
\caption{Fixed-target-patch comparison on 150 matched pairs per alias.}
\label{tab:pixel150}
\end{table}

We further examine background sensitivity using a gray-background control M for all 11 models and a same-domain shuffled scene S for eight models (Table~\ref{tab:pixel150-controls}).
Since A and B produce identical gray inputs, each target and model shares a single M response.
All eight models evaluated on S have higher RR in A than in S, with intervals entirely above zero for both 235B-A22B variants.
The gray control also reveals that removing the scene can increase rewriting: RR reaches 25.33\% for 8B Instruct and 36.67\% for 235B-A22B Instruct, exceeding both corresponding full-scene rates.

\begin{table}[htbp]
\centering
\small
\setlength{\tabcolsep}{4pt}

\begin{tabular}{lrrrrrrr}
\toprule
Model alias & A: O & B: O & M: L & M: C & S: C & $\Delta_{A-S}$ & 95\% CI \\
\midrule
GPT-5.2 & 2.00 & 3.33 & 90.00 & 9.33 & 12.00 & $+5.33$ & $[-1.33, 12.00]$ \\
GPT-5.5 & 3.33 & 2.00 & 88.67 & 10.00 & 10.00 & $+6.00$ & $[-0.67, 12.67]$ \\
Claude Sonnet 4.6 & 0.00 & 1.33 & 99.33 & 0.67 & 0.00 & $+0.67$ & $[0.00, 2.00]$ \\
Gemini 3.1 Flash Lite & 0.00 & 0.00 & 98.00 & 1.33 & -- & -- & -- \\
Gemini 3.5 Flash & 0.67 & 2.00 & 95.33 & 1.33 & 2.00 & $+0.67$ & $[0.00, 2.00]$ \\
Qwen3-VL-8B-I & 4.00 & 3.33 & 66.67 & 25.33 & 7.33 & $+0.67$ & $[-3.35, 4.67]$ \\
Qwen3-VL-8B-T & 8.00 & 8.67 & 81.33 & 12.00 & -- & -- & -- \\
Qwen3-VL-32B-I & 5.33 & 3.33 & 92.00 & 6.00 & 10.00 & $+2.67$ & $[-1.33, 6.67]$ \\
Qwen3-VL-32B-T & 6.67 & 7.33 & 88.67 & 8.00 & -- & -- & -- \\
Qwen3-VL-235B-A22B-I & 4.00 & 2.67 & 58.67 & 36.67 & 10.00 & $+6.67$ & $[1.33, 12.00]$ \\
Qwen3-VL-235B-A22B-T & 4.00 & 4.00 & 88.00 & 6.00 & 6.00 & $+4.00$ & $[1.33, 7.33]$ \\
\bottomrule
\end{tabular}
\caption{Additional outcomes on the same 150 targets. }
\label{tab:pixel150-controls}
\end{table}

Together, these comparisons establish sensitivity to surrounding visual input under fixed target pixels.
The A--B comparison captures the combined influence of scene meaning, identifier role, and visual appearance, while the gray control tests the effect of removing the surrounding scene.
\section{Consensus Filtering and Recoverable Shared Errors}
\label{app:consensus}

The archived three-model panel used GPT-5.2, Claude Sonnet 4.6, and gemini-3-flash.
Exact agreement accepted 474 of 781 images, including 18 nonliteral answers (16 C, two O; Table~\ref{tab:consensus-selection}).
Adding Qwen3-VL-32B Thinking kept all 18 errors but removed 165 literal answers.

\begin{table}[htbp]
\centering\small

\setlength{\tabcolsep}{5pt}
\renewcommand{\arraystretch}{1.12}
\begin{tabularx}{\linewidth}{@{}Xrrrr@{}}
\toprule
Agreement rule & Accepted $n$ & Coverage (\%) & L/C/O & Error (\%) \\
\midrule
Three models, exact & 474 & 60.69 & 456/16/2 & 3.80 \\
Four models, exact & 309 & 39.56 & 291/16/2 & 5.83 \\
Three models, normalized & 493 & 63.12 & 465/26/2 & 5.68 \\
Four models, normalized & 327 & 41.87 & 299/26/2 & 8.56 \\
\bottomrule
\end{tabularx}
\caption{Consensus filtering on the archived 781-image evaluation.}
\label{tab:consensus-selection}
\end{table}



\section{Crossing Target Blur with Peripheral Scene Removal}
\label{app:blur-mask}

Four models have clear/moderate $\times$ full/gray responses for the same 781 images (12,496 outputs).
Gray replaces pixels outside the red box with RGB $(128,128,128)$; moderate blur uses the fixed recipe in Section~\ref{sec:blur}.
At each clarity level, full and gray images retain identical target pixels, position, and scale.
Table~\ref{tab:blur-mask-lco781} reports all L/C/O rates.

\begin{table}[H]
\centering
\begingroup
\small\fontencoding{OT1}\fontfamily{ptm}\selectfont
\renewcommand{\arraystretch}{1.16}
\setlength{\tabcolsep}{2pt}

\begin{tabular*}{\linewidth}{@{\extracolsep{\fill}}l*{12}{r}@{}}
\toprule
\multirow{2}{*}{\textbf{Model alias}}
& \multicolumn{3}{c}{\textbf{Full clear}}
& \multicolumn{3}{c}{\textbf{Full moderate}}
& \multicolumn{3}{c}{\textbf{Gray clear}}
& \multicolumn{3}{c}{\textbf{Gray moderate}} \\
\cmidrule(lr){2-4}
\cmidrule(lr){5-7}
\cmidrule(lr){8-10}
\cmidrule(l){11-13}
& L & C & O & L & C & O & L & C & O & L & C & O \\
\midrule
GPT-5.2
& 67.35 & 29.58 & 3.07
& 50.19 & 45.33 & 4.48
& 84.25 & 11.01 & 4.74
& 68.25 & 20.87 & 10.88 \\
Gemini 3.5 Flash
& 89.76 & 8.96 & 1.28
& 80.41 & 18.44 & 1.15
& 96.93 & 2.43 & 0.64
& 87.32 & 10.76 & 1.92 \\
Claude Sonnet 4.6
& 85.02 & 12.16 & 2.82
& 71.06 & 22.66 & 6.27
& 96.80 & 2.56 & 0.64
& 82.71 & 11.40 & 5.89 \\
Qwen3-VL-32B T.
& 47.50 & 44.43 & 8.07
& 35.60 & 51.47 & 12.93
& 67.86 & 24.33 & 7.81
& 58.64 & 25.61 & 15.75 \\
\bottomrule
\end{tabular*}
\endgroup

\caption{Transcription outcomes across blur and scene-removal conditions.
All rates are percentages. T.\ denotes Thinking.}
\label{tab:blur-mask-lco781}
\end{table}

For each outcome, the blur effect is the moderate-minus-clear rate; the interaction is the full blur effect minus the gray blur effect.
Pointwise intervals use 20,000 paired image-bootstrap resamples.
All literal-accuracy interaction intervals include zero, while canonical-rewriting interactions are positive for GPT and Qwen.
Together, these results suggest that surrounding cues can reinforce canonical rewriting when visual evidence from the target becomes weaker.

\end{document}